\documentclass[conference]{IEEEtran}
\IEEEoverridecommandlockouts
\usepackage{cite}
\usepackage{amsmath,amssymb,amsfonts}
\usepackage{algorithmic}
\usepackage{graphicx}
\usepackage{textcomp}
\usepackage{xcolor}
\usepackage{subcaption}
\usepackage{hyperref}
\usepackage{float}
\usepackage{multirow}
\usepackage{stfloats}
\usepackage{flushend}

\def\BibTeX{{\rm B\kern-.05em{\sc i\kern-.025em b}\kern-.08em
    T\kern-.1667em\lower.7ex\hbox{E}\kern-.125emX}}
\begin{document}

\title{GeoNLI - A Natural Language Interpreter for Satellite Imagery\\[0.4em]
}


\author{
  Ashutosh Gandhe${^1}$,
  Anupam Rawat${^1}$,
  Geet Sethi${^1}$,
  Kabir Nasiruddin${^1}$,\\
  Madhav Kotecha${^1}$,
  Panav Shah${^1}$,
  Rakshit Sawarn${^1}$,
  Soumitra Nayak${^1}$ \\
  ${^1}$Indian Institute of Technology, Bombay\\
}

\maketitle

\begin{abstract}
Multi-modal multitasking models have shown strong performance on remote sensing datasets. However, because these models are trained on heterogeneous data and vary across tasks, designing a unified model that performs well in captioning, visual question answering (VQA), and visual grounding remains challenging.

In this work, we evaluate several models on the VRS Bench and NWPU-VHR-10 datasets. The \textit{EarthMind} model demonstrates strong results in both captioning and VQA. For grounding, we propose multiple pipelines---\textit{RemoteSAM-SAM-v1}, \textit{RemoteSAM-SAM-v2}, and \textit{DiffuSAM}---and ultimately adopt a majority-voting ensemble across \textit{EarthMind}, \textit{RemoteSAM}, \textit{SAM3}, \textit{Falcon}, \textit{RemoteSAM-SAM3-v1}, \textit{RemoteSAM-SAM3-v2}, and \textit{DiffuSAM} predictions.

Our unified, modular pipeline integrates advanced SAM variants with multimodal LLMs to jointly perform captioning, VQA, and grounding. It achieves 82\% accuracy on captioning and 83.32\% on VQA, with 90.94\%, 52.04\%, and 92.06\% for binary, numeric, and semantic question types respectively. For grounding, it attains 64.94\% accuracy.

By combining diverse VLMs with our custom \textit{RemoteSAM--SAM3} models through ensemble majority voting, the system delivers more accurate and consistent remote-sensing understanding than task-specific approaches.



\end{abstract}


\section{\textbf{Introduction}}
\label{sec:intro}

Remote sensing (RS) imagery plays a crucial role in applications such as urban planning, national security, and environmental monitoring. With the rapid growth of satellite data, deep learning–based methods have become essential for automating RS tasks. Advances in sensor technology have further enhanced data quality, enabling increasingly capable models.


Multitask VLMs generally adopt either (1) shared encoders with task-specific heads or (2) prompt-based unification, where the model interprets natural-language instructions (e.g., “Describe the image” or “Locate the object referred to as …”). Modern VLMs typically follow the second paradigm due to reduced parameter overhead and greater flexibility across tasks.


Multitask learning offers substantial benefits: shared representations allow tasks to reinforce one another—segmentation aids grounding, detection supports VQA, and language guidance improves spatial reasoning. A unified system also simplifies deployment in real-world geospatial applications requiring comprehensive scene understanding.

Although many VLMs exist, they are trained on heterogeneous datasets and evaluated under inconsistent settings. To establish a standardized and effective pipeline, we evaluate several models and propose a unified framework. We use the EarthMind model for captioning and VQA, as both tasks rely on text generation from images. \textbf{EarthMind excels here due to its RoI-based summarization and self-instruct VQA training, but it performs poorly on grounding.} 

\begin{figure}[h!]
\centering
\includegraphics[width=0.24\textwidth]{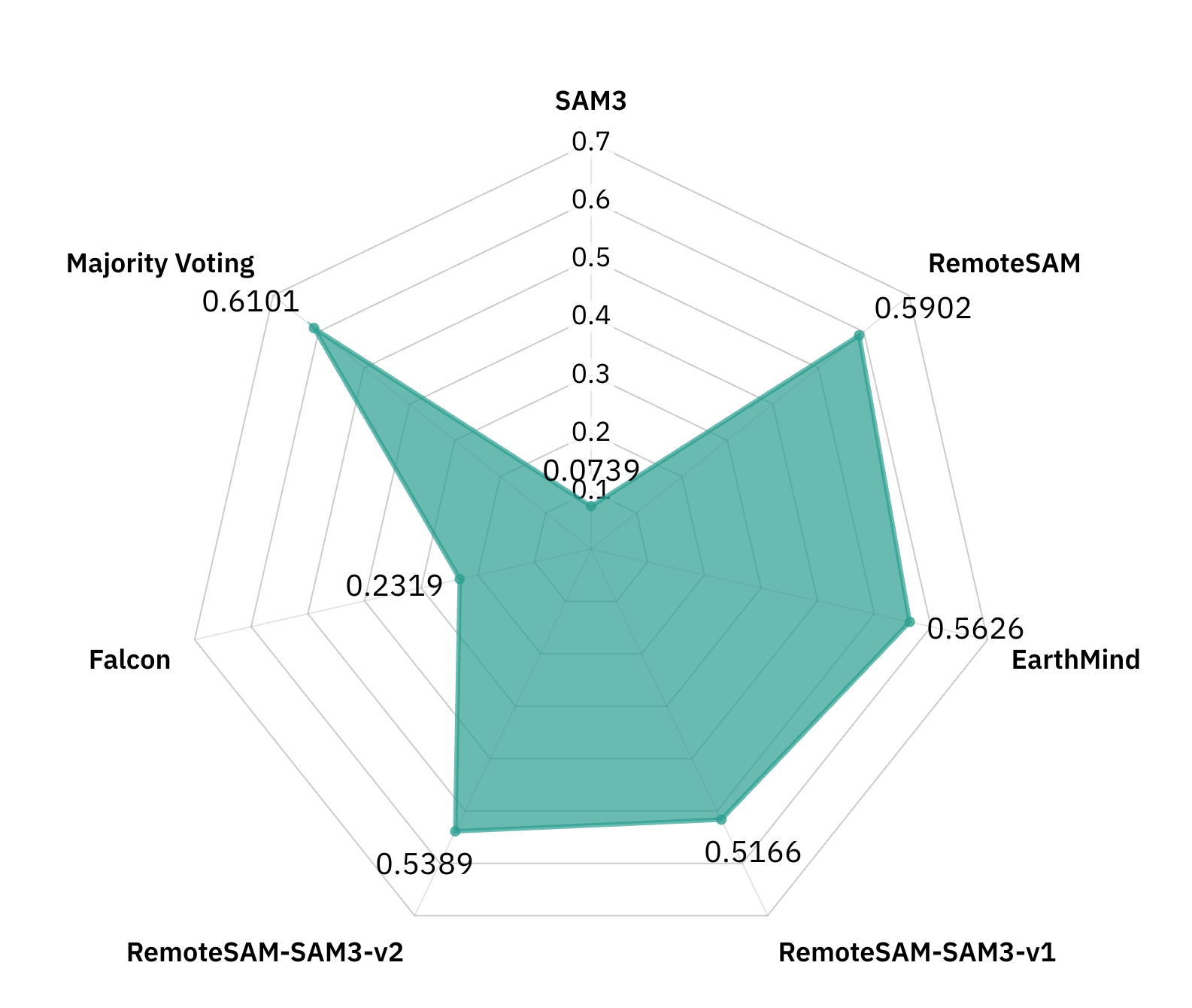}
\includegraphics[width=0.24\textwidth]{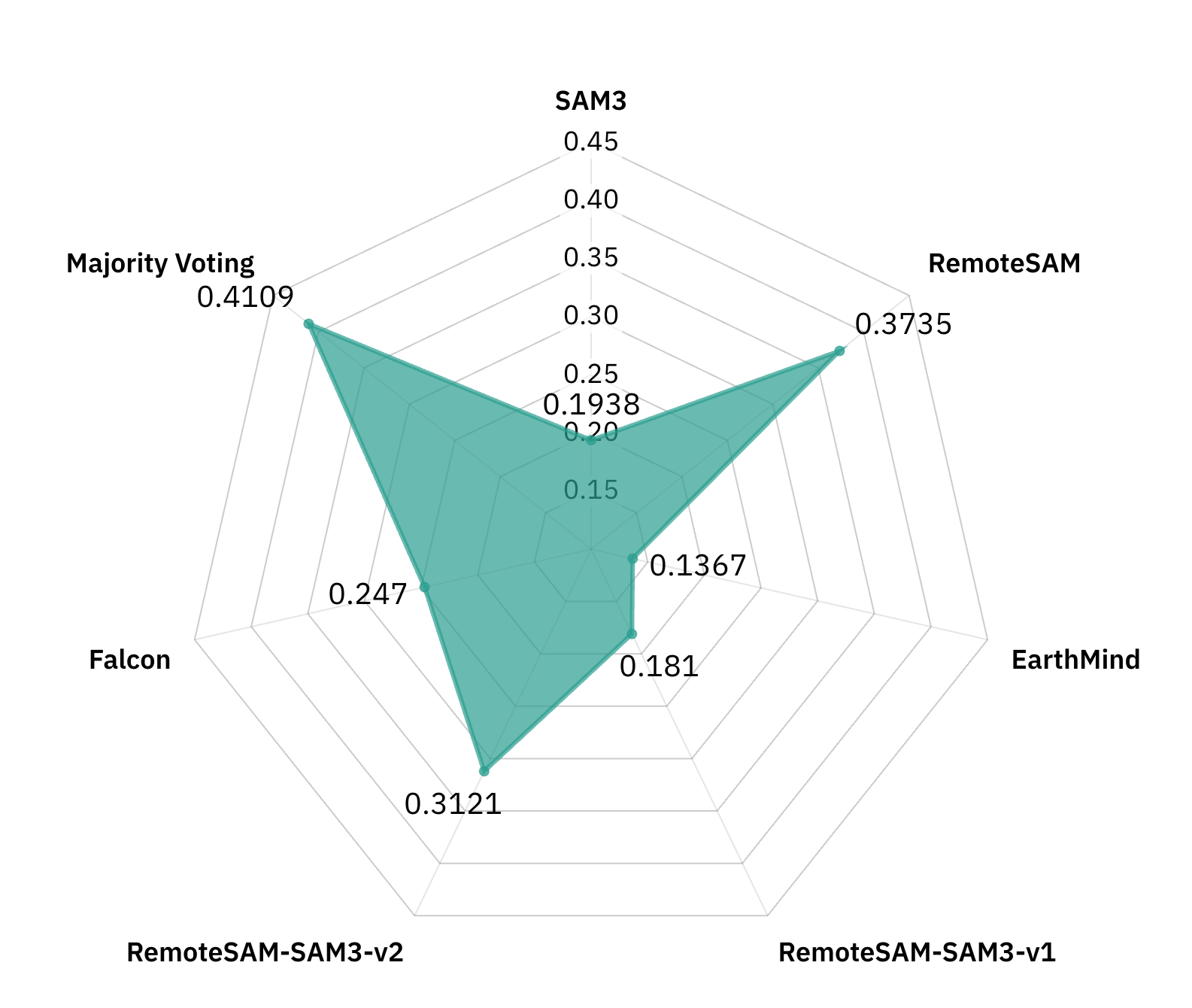}
\caption{Scores of several foundational models for grounding}
\end{figure}

\begin{figure*}[!t]
\centering
\includegraphics[width=0.23\textwidth]{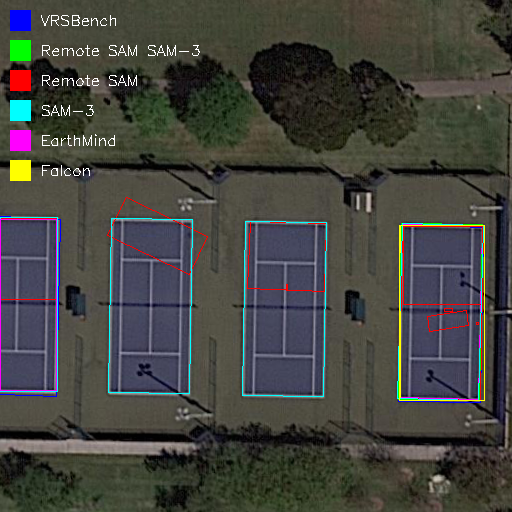}
\includegraphics[width=0.23\textwidth]{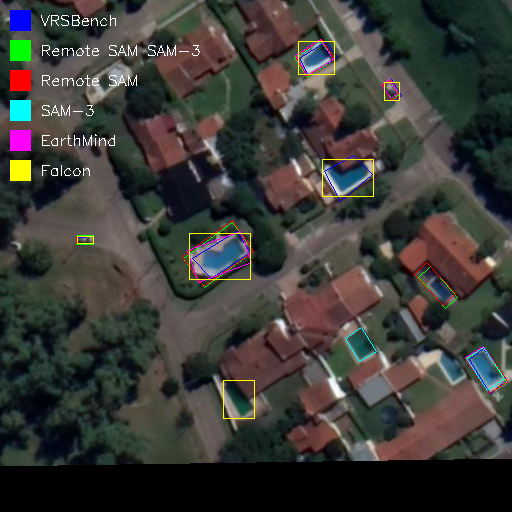}
\includegraphics[width=0.23\textwidth]{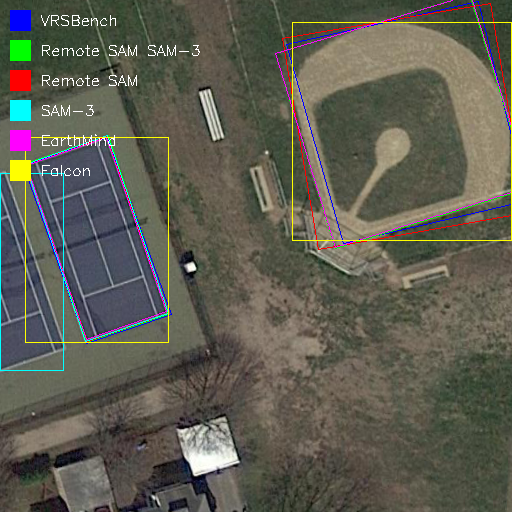}
\includegraphics[width=0.23\textwidth]{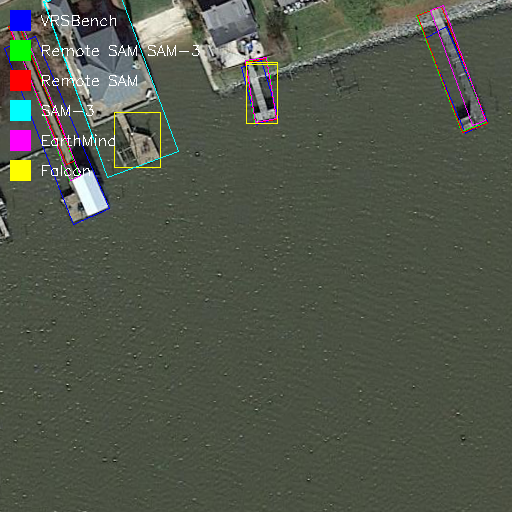}
\caption{Outputs of various models}
\end{figure*}

\section{\textbf{Innovation \& Novelty}}
\begin{itemize}
    \item We have designed three grounding pipelines, RemoteSAM-SAM3-v1, RemoteSAM-SAM3-v2 and DiffuSAM, for grounding task. We report the performance of both the individual components of each model and their majority-voting ensembles.
    \item We propose an unified pipeline for captioning, visual question answer, and visual grounding. For captioning and VQA, we use the EarthMind model, running it in inference mode for captioning, and finetuning it on an augmented dataset for VQA. For the grounding task, we have employed a combination of EarthMind, RemoteSAM, SAM3 and Falcon.\\
    We have also developed three new model pipelines RemoteSAM-SAM3-v1, RemoteSAM-SAM3-v2 and DiffuSAM.
    
    \item We have done a thorough experimentation on two datasets, VRS Bench and NWPU-VHR-10 on the unified pipeline as well as the individual and existing models. The results show that the unified pipeline produces the best outputs compared to the existing models.
    \item Our model’s USP lies in its unified and modular design that brings together advanced SAM variants, and multimodal LLMs to handle captioning, VQA, and grounding within a single coherent pipeline. By combining EarthMind with other VLMs, our custom RemoteSAM-SAM3 models and using ensemble majority voting, the system achieves more accurate, consistent, and reliable remote-sensing understanding than existing task-specific approaches.
\end{itemize}

The structure of this manuscript is as follows: Section \ref{sec:problem} shows the detailed problem statement, \ref{sec:related} presents the literature survey, \ref{sec:methodology} presents the proposed method. Sections  \ref{sec:implemetation} and \ref{sec:results} depict the implementation details and results. Section \ref{sec:challenges} presents the challenges and future works and finally section \ref{sec:conclusion} concludes this work.

\section{\textbf{Detailed Problem Description}}
\label{sec:problem} 

The aim of this work is to implement a multimodal multitask model for remote sensing data. The tasks are as follows:

\begin{itemize}
    \item \textbf{Captioning:} For each image, the model generates a caption that summarizes the key elements of the satellite scene. The input for this task is an image, and the output is a text description.

\item \textbf{Visual Question Answering (VQA):} VQA involves answering a question based on the given image. Questions may require geometric or semantic understanding, and the answers may be binary (``yes''/``no''), numeric, or semantic (string). The inputs are an image and a text question, and the output is a textual answer.

\item \textbf{Visual Grounding:} Visual grounding refers to localizing objects in an image based on a language query. The inputs are an image and a text phrase referring to the target object, and the output is an image with bounding box coordinates. Depending on the model design, the system may alternatively return only the bounding box coordinates instead of a grounded image.
    
\end{itemize}

\section{\textbf{Related Works}}
\label{sec:related}

Research in multimodal remote sensing has been enabled primarily by the emergence of large-scale image–text datasets that support captioning, VQA, grounding, and high-resolution semantic understanding etc. Early collections such as RSICD \cite{lu2017exploring}, UCM-Captions \cite{Ali2018}, and Sydney-Captions \cite{qu2016deep} introduced paired aerial imagery and natural-language descriptions, while BigEarthNet \cite{clasen2024reben} and RSVG \cite{zhan2023rsvg} expanded the scope toward grounding and fine-grained semantics. More recent benchmarks, including RS5M \cite{10679571} for CLIP-style pretraining, VRSBench \cite{li2024vrsbench} for multitask evaluation, XLRS-Bench \cite{wang2025xlrs} for extremely high-resolution imagery, and multiple instruction datasets provide the scale and diversity needed for developing modern vision–language models in the remote sensing domain.

General task-specific models such as SAM \cite{kirillov2023segany}, SAM2 \cite{ravi2024sam2}, SAM3 \cite{carion2025sam}, GroundingDINO \cite{liu2023grounding}, BLIP \cite{li2022blip}, and GLIP \cite{li2021grounded} have served as foundational architectures for segmentation, grounding, and captioning across a broad range of imagery. These models, although not RS-specific, are frequently adapted as backbones or initialization points for remote sensing applications. Correspondingly, RS-specific task models, such as RemoteSAM \cite{yao2025remotesam}, RSVG \cite{zhan2023rsvg} etc. incorporate multi-scale feature extractors and remote sensing specific data prior to address challenges unique to aerial and satellite imagery, including dense object distributions, ultra-high resolution, multispectral variation, and fine-grained land-use semantics.
\begin{figure*}[t]
    \centering
    \includegraphics[width=0.75\linewidth]{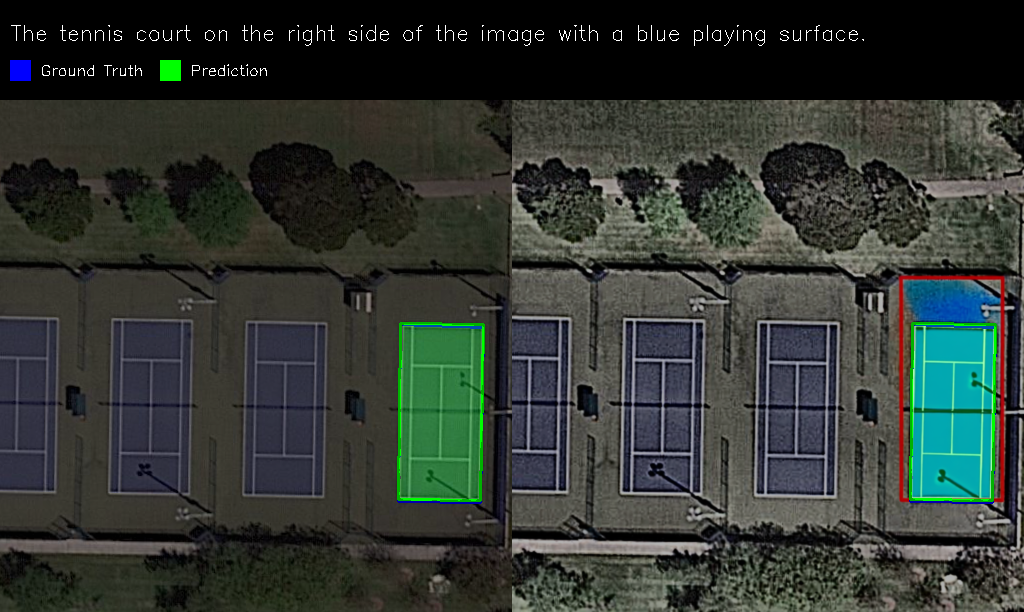}
    \caption{Image generated by DiffuSAM (Red box - generated by Qwen-Image-Edit, Green box - Final bounding box)}
    \label{pipeline}
\end{figure*}
Building on this foundation, multitask VLMs such as LLaVA \cite{liu2023llava}, Qwen-VL \cite{Qwen-VL}, BLIP-2 \cite{li2023blip}, Flamingo \cite{alayrac2022flamingo}, and MiniGPT-4 \cite{zhu2023minigpt} provide unified pipelines for captioning, VQA, grounding, and instruction following, and have inspired a new generation of remote sensing–oriented multitask models. RS-specific systems like Falcon \cite{yao2025falcon}, EarthMind \cite{shu2025earthmindleveragingcrosssensordata}, RSGPT \cite{hu2025rsgpt}, GeoGPT \cite{zhang2024geogpt}, GeoChat \cite{kuckreja2023geochat}, EarthGPT \cite{zhang2024earthgpt} extend these architectures with geospatial pretraining, RS instruction tuning, multi-scale processing for large images, and sensor-aware visual encoders. These models, trained on datasets like RS5M and VRSBench, act as general-purpose multimodal VLMs capable of supporting captioning, spatial reasoning, grounding, and high-resolution scene interpretation within a single integrated framework.



\section{\textbf{Methodology}}
\label{sec:methodology}

Since existing models are trained and evaluated on different datasets, comparing their performance fairly is challenging. To address this, we evaluate all models on a common benchmark. Because model performance also varies significantly across tasks, we assess captioning, VQA, and grounding separately. Our results show that EarthMind performs strongly on captioning and VQA. Its RoI-based summarization and few-shot self-instruction prompting contribute substantially to this performance. However, EarthMind struggles with grounding, which is more challenging and fundamentally different from the other two tasks.

To improve grounding performance, we employ multiple pipelines—including EarthMind, RemoteSAM, SAM3, Falcon, RemoteSAM-SAM3-v1, RemoteSAM-SAM3-v2, and DiffuSAM—and apply a majority voting strategy. Details of the proposed models are provided below, while descriptions of existing models are included in the Appendix.\ref{appendix_models}.



\begin{figure*}[t]
    \centering
    \begin{subfigure}{0.45\linewidth}
        \centering
    \includegraphics[width=\linewidth]{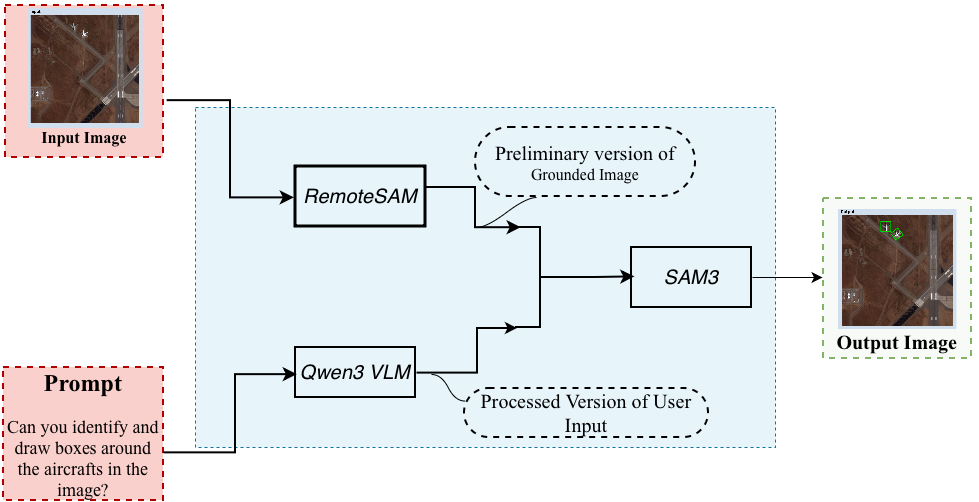}
        \caption{\textbf{RemoteSAM-SAM3-v1}}
        \label{remotesam_sam3_v1}
    \end{subfigure}
    \hfill
    \centering
    \begin{subfigure}{0.45\linewidth}
        \centering
    \includegraphics[ width=\linewidth]{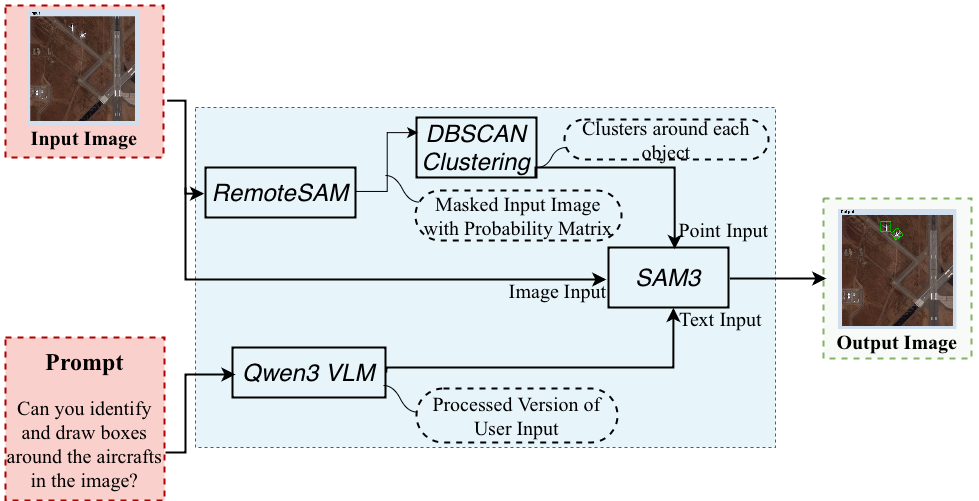}
        \caption{\textbf{RemoteSAM-SAM3-v2}}
        \label{remotesam_sam3_v2}
    \end{subfigure}
    \hfill
    \begin{subfigure}{0.70\linewidth}
        \centering
    \includegraphics[width=\linewidth]{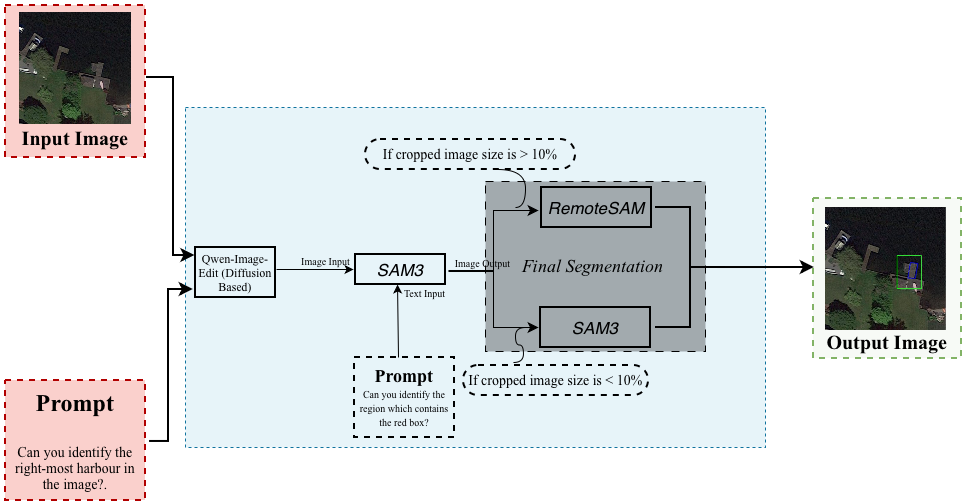}
        \caption{\textbf{DiffuSAM}}
        \label{Diffusion}
    \end{subfigure}

    \caption{The architectures of the components of the unified pipeline (proposed). a) RemoteSAM-SAM3-v1 Architecture, b) RemoteSAM-SAM3-v2 Architecture, c) DiffuSAM Architecture.}
    \label{architectures2}
\end{figure*}

\subsection{\textbf{RemoteSAM-SAM3}}
We propose a combined pipeline that integrates RemoteSAM and SAM3. Since SAM3 is capable of general-purpose grounding and RemoteSAM is specifically trained on remote sensing data, the two models complement each other, motivating the design of a unified framework that leverages the strengths of both.
\begin{enumerate}
    \item \textbf{RemoteSAM-SAM3-v1}: In this version, the input image and text prompt are passed to both the RemoteSAM model and a large language model (Qwen). RemoteSAM produces an initial grounded output, while Qwen identifies the target object for grounding. The detected object and the grounded output from RemoteSAM are then provided to SAM3 as a text prompt and a box prompt, respectively, to generate the final grounding result. However, because the grounding output from RemoteSAM is often fragmented and not spatially continuous, SAM3 may still struggle with accurate grounding in this setting.

    \item \textbf{RemoteSAM-SAM3-v2:} To address the limitations of the previous model, we design an improved version that combines RemoteSAM and SAM3 to produce more reliable grounded object segmentation. The image and text prompt are first passed through RemoteSAM to obtain a logit mask. Since these masks often contain fragmented regions, we apply DBSCAN clustering to group them into coherent object clusters, resulting in more spatially continuous grounding candidates.

Next, we extract object names from the prompt using a large language model (Qwen). 
The image, object name, and cluster mask are then jointly provided to SAM3.

Because an image may contain multiple objects, SAM3 is executed separately for each cluster, producing several candidate grounded outputs. These candidates are evaluated by computing their IoU with the associated DBSCAN cluster. The candidate with the highest IoU is selected as the final grounding result for that object. Repeating this process across all clusters yields a final output in which each object is grounded accurately, combining the semantic correctness of the LLM with the spatial precision of clustering.

\end{enumerate}

Figure \ref{remotesam_sam3_v1} and \ref{remotesam_sam3_v2} show the two versions of this architecture.

\subsection{\textbf{DiffuSAM}}
The core idea of this model is to use a diffusion-based approach (Qwen-Image-Edit) that, given a modified text prompt, generates an image containing a red bounding box around the region of interest. This annotated image is then passed to the SAM3 model, which produces a mask within the red box; the mask is used to compute an approximate bounding box. The cropped image corresponding to this box is subsequently forwarded to either RemoteSAM or SAM3.

Our experiments show that RemoteSAM performs better than SAM3 when there are mutliple objects to find the target object from, likely because RemoteSAM is trained specifically on visual grounding. For small cropped areas, however, SAM3 achieves superior performance, likely because the smaller cropped regions are more likely to only contain the required object (i.e. the inital approximate bounding box was fairly accurate). Therefore, if the cropped region covers more than 10\% of the image, we use RemoteSAM; otherwise, we use SAM3. Directional keywords such as ``left'' or ``right'' are removed from the prompt at this stage, since the spatial information is already encoded in the cropped image and retaining these words can introduce confusion.

The final mask obtained from RemoteSAM or SAM3 is then used to compute the final bounding box. Figure~\ref{Diffusion} illustrates the architecture of this model.

\begin{figure*}[t]
    \centering
    \includegraphics[width=0.75\linewidth]{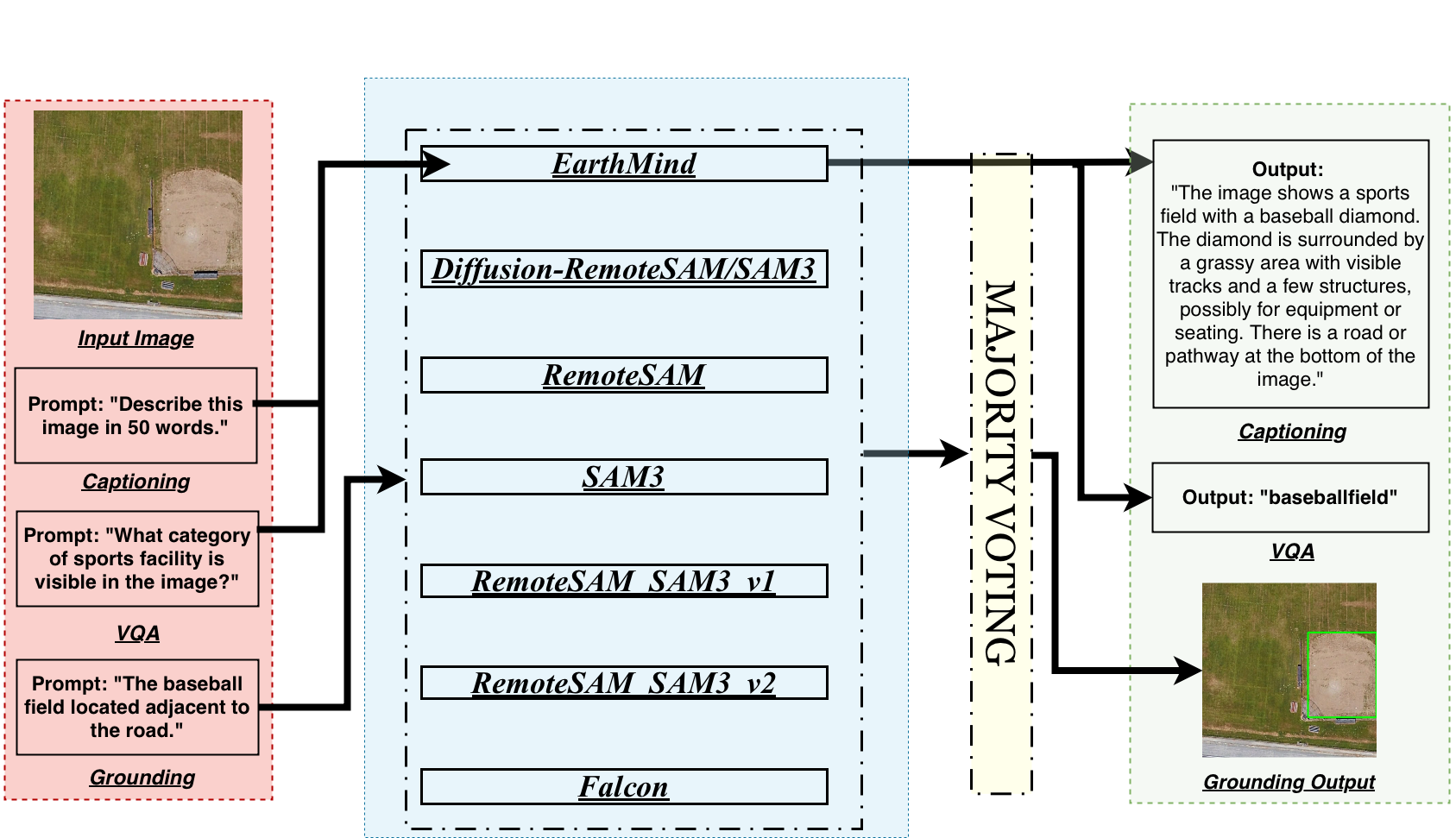}
    \caption{The overall architecture of the proposed pipeline is illustrated here. For captioning and VQA tasks, we use the EarthMind model, whereas for the grounding task, we use a majority-voting ensemble of EarthMind, RemoteSAM, SAM3, Falcon, RemoteSAM-SAM3-v1, RemoteSAM-SAM3-v2, and DiffuSAM.}
    \label{pipeline}
\end{figure*}

\subsection{\textbf{Overall Architecture}}

Since EarthMind performs well on VQA and captioning tasks, we use it for both. For captioning, inference is performed directly using the pre-trained model. Initial results reveal a weakness in numeric VQA, indicating that greater emphasis is needed on object-counting accuracy during supervised fine-tuning (SFT). To address this, the final SFT dataset is reweighted with 65\% numeric, 20\% semantic, and 15\% binary questions. Additionally, to preserve EarthMind's ability to perform all tasks it was originally trained on and to prevent catastrophic forgetting, we include a subset of datapoints from its original training dataset when constructing the final SFT dataset.

For the grounding task, we use the following models: EarthMind, RemoteSAM, SAM3, Falcon, RemoteSAM-SAM3-v1, RemoteSAM-SAM3-v2, and DiffuSAM. We generate grounding outputs from all models and apply a majority voting strategy to select the best prediction. This ensures robust performance across tasks within a unified pipeline. Equation~\ref{majority} presents the majority voting formulation, where $score[i]$ denotes the overlap between the prediction of model $i$ and those of the remaining models. The prediction with the highest agreement is selected as the final output. Figure~\ref{pipeline} illustrates the overall pipeline.

\begin{equation}
\begin{aligned}
\text{score}[i] &= \sum_{j \neq i} \exp\!\left(-\,0.5|\text{boxes}(i) - \text{boxes}(j)|\right)\cdot \text{mIoU}(i,j), \\
i^{*} &= \arg\max_i \ \text{score}[i]
\end{aligned}
\label{majority}
\end{equation}

The exponential term serves as a penalty applied when the number of predicted outputs differs across models.

\subsection{\textbf{SAR-to-Optical Image translation}}

To improve the interpretability of Synthetic Aperture Radar (SAR) data within our unified remote sensing pipeline, we used a conditional diffusion model that translates SAR images into corresponding optical imagery. The details of the model is provided in Appendix \ref{appendix_models_sar}.

In order to detect if the image is SAR or optical image, we have used the following conditions. First, we calculated mean absolute difference between Blue, Green, and Red channels. If the difference is less than 2, we use tag it as SAR. Next, we checked `Speckle-Denoising' Ratio, where, the presence of granular speckle noise is checked using $\text{Ratio} = \frac{\sigma_{orig}}{\sigma_{blur}}$. If the ratio is greater than 1.15, it is SAR image. Next, we checked the cofficient of variation by computing $CV = \frac{\sigma}{\mu}$. If it is greater than 0.45, it is SAR image. Then, Effective Number of Looks (ENL) is computed using $\text{ENL} = \frac{\mu^2}{\sigma^2}$. If it is less than 8, the image is SAR. Then, the skewness and kurtosis of the histogram is checked. If skewness is greater than 1 and kurtosis is greater than 2, it is denoted as SAR image. If all the conditions fail, it is considered as optical image. Since we apply multiple conditions, there is very low probability that a optical image is considered as SAR image.

\section{\textbf{Implementation Details}}
\label{sec:implemetation}

\subsection{\textbf{Experimental Setup}}

Two NVIDIA A100 GPUs with 80\,GB of memory each are used for all experiments, and all scripts are implemented using the PyTorch framework. We perform evaluation in inference mode for both captioning and grounding tasks. For the VQA task, however, we observe a clear performance gain after fine-tuning the EarthMind model; therefore, EarthMind is fine-tuned specifically for VQA.

\subsection{\textbf{Dataset Details}}
\label{sec:data}

\begin{itemize}
    \item \textbf{VRS Bench \cite{li2024vrsbench}:} VRS Bench is a large-scale, multi-task vision–language benchmark for remote sensing which can support captioning, VQA and visual grounding tasks. It contains 29614 images, 29614 human-verified captions, 52472 referring-expression annotations, and 123221 VQA pairs. Each image has resolution $512 \times 512$.  It is built through a semi-automatic pipeline combining object-level metadata extraction, LLM-based annotation generation, and human verification, resulting in high-quality, fine-grained annotations. 
    \item \textbf{NWPU-VHR-10 \cite{su2019object}:} NWPU-VHR-10 is a very-high-resolution remote sensing object-detection dataset designed for evaluating detection algorithms in complex urban and industrial scenes. It contains 800 images, including 650 positive images with 3651 annotated objects and 150 negative images. The dataset covers 10 categories - airplane, ship, storage tank, baseball diamond, tennis court, basketball court, ground track field, harbor, bridge, and vehicle. Images have spatial resolutions ranging from 0.2 m to 1.0 m, with varying image sizes. All instances are annotated using axis-aligned bounding boxes, providing high-quality labels for benchmarking small-object detection and multi-scale recognition in VHR remote sensing imagery.
\end{itemize}

\subsection{\textbf{Evaluation Details}}
\label{sec:eval}

To comprehensively assess captioning, question answering, and grounding performance, we use task-specific metrics. For captioning, we use the BERT-BLEU$_4$ score ($S_{\text{captioning}}$). For VQA, we evaluate binary attribute accuracy ($S_{\text{binary}}$), numeric attribute score ($S_{\text{numeric}}$), and semantic attribute score ($S_{\text{semantic}}$). For grounding, we compute the mIoU with a count penalty ($S_{\text{grounding}}$). Detailed definitions of these metrics are provided in Appendix~\ref{appendix_metrics}.

For the final evaluation of the overall pipeline, we use the following weighted score:
\begin{equation}
\begin{aligned}
\text{Final Score} ={}&
0.20\,S_{\text{captioning}} +
0.30\,S_{\text{grounding}} \\
&{}+ 0.10\,S_{\text{binary}} +
0.20\,S_{\text{numeric}} +
0.20\,S_{\text{semantic}}.
\end{aligned}
\end{equation}

\subsection{\textbf{Front-End}}

The frontend of the application is built in React and serves as the primary interface for authentication, image uploads, and analysis workflows. It communicates with the FastAPI backend through REST endpoints. The backend requests are proxied to enable smooth full-stack integration. For production deployment, the frontend is compiled into optimized static assets and served via Nginx.

\begin{table}[t]
\centering
\begin{tabular}{lcc}
\hline
\textbf{Metric} & \textbf{Pretrained Version} & \textbf{Finetuned Version} \\
\hline
Total QA Pairs              & 2020   & 2020   \\
Binary Samples              & 541    & 541    \\
Numeric Samples             & 426    & 426    \\
Semantic Samples            & 1053   & 1053   \\
$S_{binary}$                   & 0.9094 & 0.9094 \\
$S_{numeric}$                  & 0.4753 & 0.5204 \\
$S_{semantic}$                 & 0.9015 & 0.9206 \\
Overall Accuracy            & 0.8137 & 0.8332 \\
\hline
\end{tabular}
\caption{Comparison of pre-trained and finetuned versions (EarthMind model) on the VRS Bench Dataset for VQA task.}
\label{tab:vqa-vrs}
\end{table}

\begin{table}[t]
\centering
\begin{tabular}{lcc}
\hline
\textbf{Metric} & \textbf{Pretrained Version} & \textbf{Finetuned Version} \\
\hline
$S_{\text{binary}}$          & 0.7482 & 0.9326 \\
$S_{\text{numeric}}$         & 0.1994 & 0.7239 \\
$S_{\text{semantic}}$        & 0.6110 & 0.9746 \\
Overall Accuracy             & 0.4527 & 0.8605 \\
Final Score @50              & 23.69  & 43.30 \\
\hline
\end{tabular}
\caption{Comparison of pre-trained and finetuned versions (EarthMind model) on the RSVQA Dataset for VQA task.}
\end{table}

\begin{table*}[h!]
\centering
\begin{tabular}{p{10cm} c c c c c c c}
\hline
\textbf{Model} & \textbf{mIoU} & \textbf{$S_{grounding}$} & \textbf{Acc@0.5} & \textbf{Acc@0.7} & \textbf{Avg Count Diff}  \\
\hline
RemoteSAM & 0.6283 & 0.5902 & 0.7812 & 0.3931 & $-0.0803$  \\
SAM3 & 0.1635 & 0.0739 & 0.2051 & 0.0887 & 0.3060 \\
RemoteSAM-SAM3-v1 & 0.5671 & 0.5166 & 0.6827 & 0.3303 & $-0.1118$  \\
RemoteSAM-SAM3-v2 & 0.6215 & 0.5389 & 0.7707 & 0.3909 & $-0.2454$  \\
EarthMind & 0.5961 & 0.5626 & 0.7414 & 0.3732 & $-0.0879$  \\
Falcon & 0.5802 & 0.2319 & 0.7221 & 0.4709 & $-1.3629$  \\
RemoteSAM + SAM3 + RemoteSAM-SAM3-v1 & 0.6325 & 0.5507 & 0.7871 & 0.3936 & $-0.2277$  \\
RemoteSAM + SAM3 + RemoteSAM-SAM3-v2 & 0.6318 & 0.5389 & 0.7870 & 0.3948 & $-0.2715$ \\
RemoteSAM + SAM3 + EarthMind & 0.5431 & 0.4754 & 0.6713 & 0.3519 & $-0.2091$   \\
RemoteSAM + RemoteSAM-SAM3-v1 + EarthMind & 0.6359 & 0.5784 & 0.7958 & 0.4047 & $-0.1683$  \\
RemoteSAM + RemoteSAM-SAM3-v1 + Falcon & 0.6430 & 0.5647 & 0.7978 & 0.4217 & $-0.2380$ \\
RemoteSAM + SAM3 + Falcon & 0.6352 & 0.5309 & 0.7899 & 0.4042 & $-0.3636$\\
RemoteSAM + RemoteSAM-SAM3-v2 + EarthMind & 0.6297 & 0.5868 & 0.7823 & 0.4080 & $-0.0958$\\
RemoteSAM + SAM3 + RemoteSAM-SAM3-v1 + EarthMind & 0.6377 & 0.5489 & 0.7988 & 0.4049 & $-0.0958$   \\
RemoteSAM + SAM3 + RemoteSAM-SAM3-v2 + EarthMind & 0.6282 & 0.5433 & 0.7826 & 0.3988 & $-0.2522$  \\
\textbf{RemoteSAM + SAM3 + RemoteSAM-SAM3-v1 + RemoteSAM-SAM3-v2 + EarthMind + Falcon + DiffuSAM}   & \textbf{0.6494} & \textbf{0.6101} & \textbf{0.7928} & \textbf{0.4121} & \textbf{0.1408} \\

\hline
\end{tabular}
\caption{Evaluation of different model combinations (VRS Bench dataset). Here, `+' denotes the majority voting of the models.}
\label{tab:model_results}
\end{table*}

\begin{table*}[h]
\centering
\begin{tabular}{p{10cm} c c c c c c}
\hline
\textbf{Model} & \textbf{mIoU} & \textbf{$S_{grounding}$} & \textbf{Acc@0.5} & \textbf{Acc@0.7} & \textbf{Avg Cnt Diff} \\
\hline
RemoteSAM & 0.5662 & 0.3735 & 0.6624 & 0.2484 & $-1.4825$\\
SAM3 & 0.5009 & 0.1938 & 0.6352 & 0.1683 & $-1.2361$ \\
RemoteSAM-SAM3-v1 & 0.5719 & 0.3121 & 0.6953 & 0.2686 & $-1.4253$ \\
RemoteSAM-SAM3-v2 & 0.5098 & 0.1810 & 0.6057 & 0.1972 & 3.2265 \\
EarthMind & 0.5256 & 0.1367 & 0.6088 & 0.1844 & $-1.1916$ \\
Falcon & 0.5415 & 0.2470 & 0.6271 & 0.3051 & $1.0584$ \\
RemoteSAM + SAM3 + RemoteSAM-SAM3-v1 & 0.5755 & 0.3394 & 0.6851 & 0.2844 & $-0.5895$ \\
RemoteSAM + SAM3 + RemoteSAM-SAM3-v2 & 0.6318 & 0.5389 & 0.7870 & 0.3948 & $-0.2715$ \\
RemoteSAM + SAM3 + EarthMind & 0.5908 & 0.3316 & 0.7426 & 0.2416 & $-0.2882$ \\
RemoteSAM + RemoteSAM-SAM3-v1 + EarthMind & 0.5711 & 0.3114 & 0.6859 & 0.2634 & $-0.7101$ \\
RemoteSAM + RemoteSAM-SAM3-v1 + Falcon & 0.5755 & 0.3394 & 0.6851 & 0.2844 & $-0.5895$ \\
RemoteSAM + SAM3 + Falcon & 0.5909 & 0.3519 & 0.7335 & 0.2485 & $-0.4109$ \\
RemoteSAM + RemoteSAM-SAM3-v2 + EarthMind & 0.6297 & 0.5868 & 0.7823 & 0.4080 & $-0.0958$ \\
RemoteSAM + SAM3 + RemoteSAM-SAM3-v1 + EarthMind & 0.5928 & 0.3263 & 0.7471 & 0.2468 & $-1.1917$ \\
RemoteSAM + SAM3 + RemoteSAM-SAM3-v2 + EarthMind & 0.5924 & 0.2630 & 0.7467 & 0.2572 & $0.6744$ \\
\textbf{RemoteSAM + SAM3 + RemoteSAM-SAM3-v1 + RemoteSAM-SAM3-v2 + EarthMind + Falcon + DiffuSAM}   & \textbf{0.6031} & \textbf{0.4109} & \textbf{0.7293} & \textbf{0.3329} & \textbf{0.9842} \\ \\
\hline
\end{tabular}
\caption{Evaluation of different model combinations (VHR dataset). Here, `+' denotes the majority voting of the models}
\label{tab:model_results}
\end{table*}

\section{\textbf{Results Analysis}}
\label{sec:results}

\subsection{\textbf{Quantitative Results}}
\label{sec:quant}

For the captioning task, the pre-trained EarthMind model achieves an accuracy of \textbf{82\%}. Tables~I and II present the results for the VQA task, where EarthMind is evaluated both as a pre-trained model and as a \textbf{fine-tuned} model. While the performance on binary questions remains nearly unchanged, the accuracy on numeric and semantic questions improves notably, especially in the case of the RSVQA dataset. Tables~III and IV report the visual grounding results of our model compared with existing approaches.

\subsection{\textbf{Qualitative Results}}
\label{sec:qual}

\begin{figure*}[h!]
    \centering
    \includegraphics[width=\linewidth]{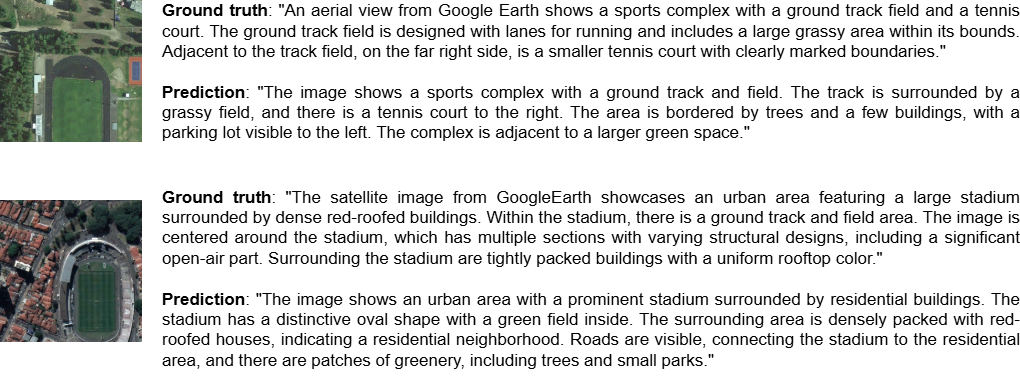}
    \caption{Qualitative results of the captioning task of EarthMind model on VRS Bench dataset.}
    \label{captioning_results}
\end{figure*}

Figure \ref{captioning_results} shows some sample outputs. From the results, we can see that the EarthMind generated captions are very good.

\begin{figure*}[h!]
    \centering
    \begin{subfigure}{\linewidth}
        \centering
    \includegraphics[width=\linewidth]{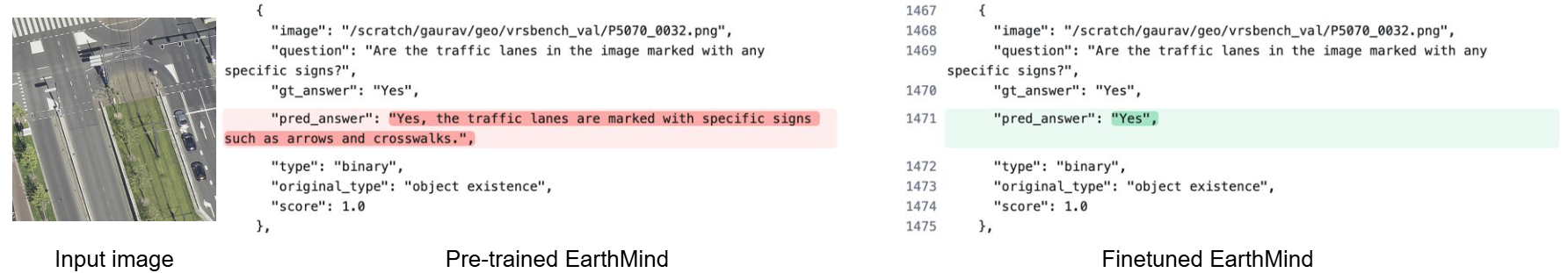}
        \caption{Binary VQA}
        \label{binary_vqa}
    \end{subfigure}
    \hfill
    \centering
    \begin{subfigure}{\linewidth}
    \includegraphics[width=\linewidth]{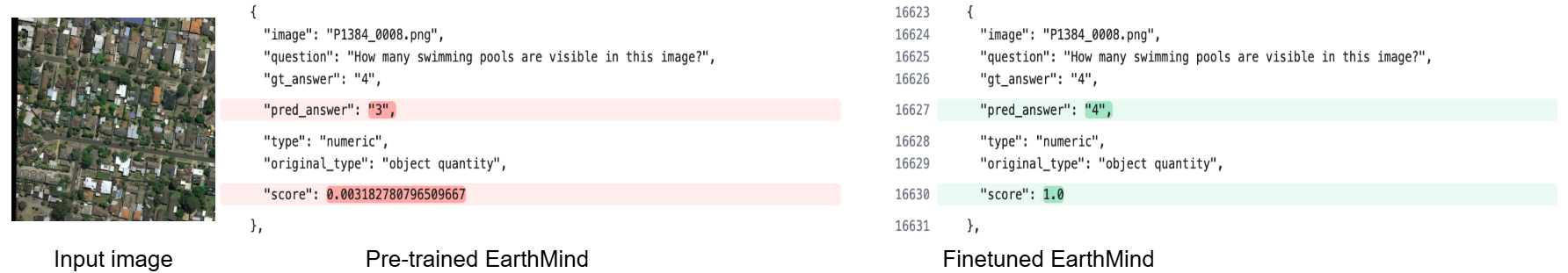}
        \caption{Numeric VQA}
        \label{numeric_vqa}
    \end{subfigure}
    \hfill
    \begin{subfigure}{\linewidth}
        \centering
    \includegraphics[width=\linewidth]{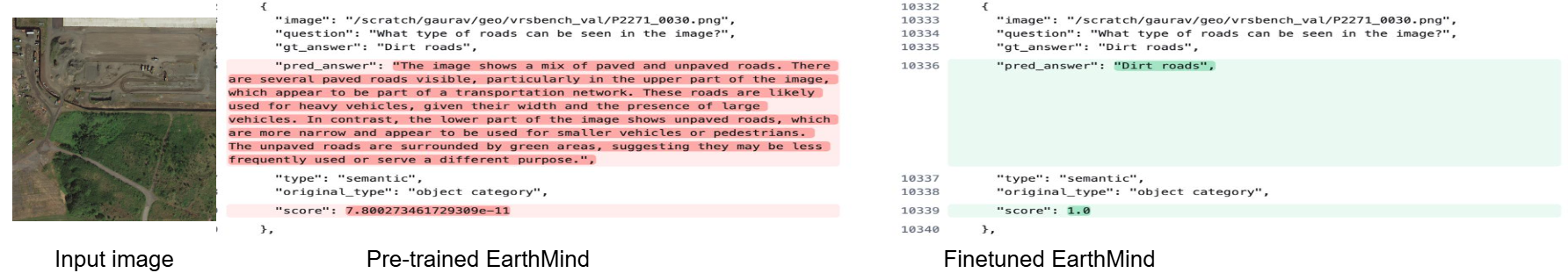}
        \caption{Semantic VQA}
        \label{semantic_vqa}
    \end{subfigure}
    \hfill

    \caption{Qualitative results of the VQA task using the pre-trained and fine-tuned EarthMind models on the VRS Bench dataset. 
(a) Binary VQA, (b) Numeric VQA, and (c) Semantic VQA. In Binary and Semantic VQA, some answers appear shortened due to instruction tuning. For Numeric and Semantic VQA, the fine-tuned model produces correct answers, whereas the pre-trained model does not.}

    \label{vqa_results}
\end{figure*}

Figure \ref{vqa_results} shows the input images, pre-trained and finetuned EarthMind model outputs for VQA task on VRS Bench dataset. For Binary task, we see that the outputs for both models are correct, but the finetuned model produces shorter output due to instruction tuning. Similarly, for semantic questions, the output is shortened. For counting and semantic questions, we see that finetuned model gives the correct numerical value, which is given incorrect in pretrained model.

\begin{figure*}[h!]
    \centering
    \begin{subfigure}{\linewidth}
        \centering
        \includegraphics[width=0.9\linewidth]{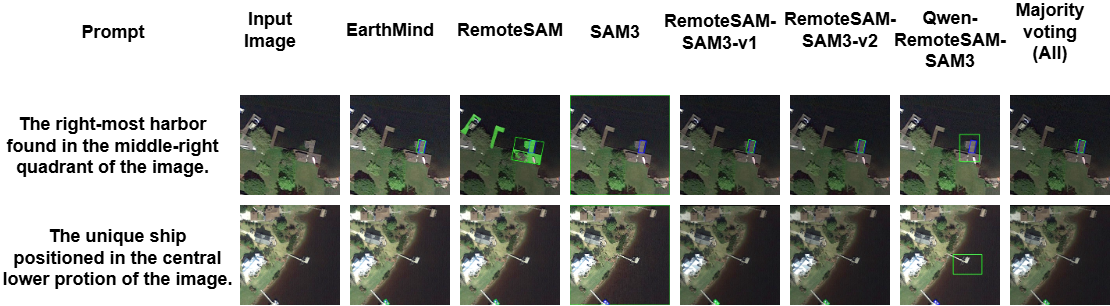}
        \caption{Qualitative results}
        \label{grounding_results_main}
    \end{subfigure}
    \hfill
    \centering
    \begin{subfigure}{0.15\linewidth}
        \includegraphics[width=\linewidth]{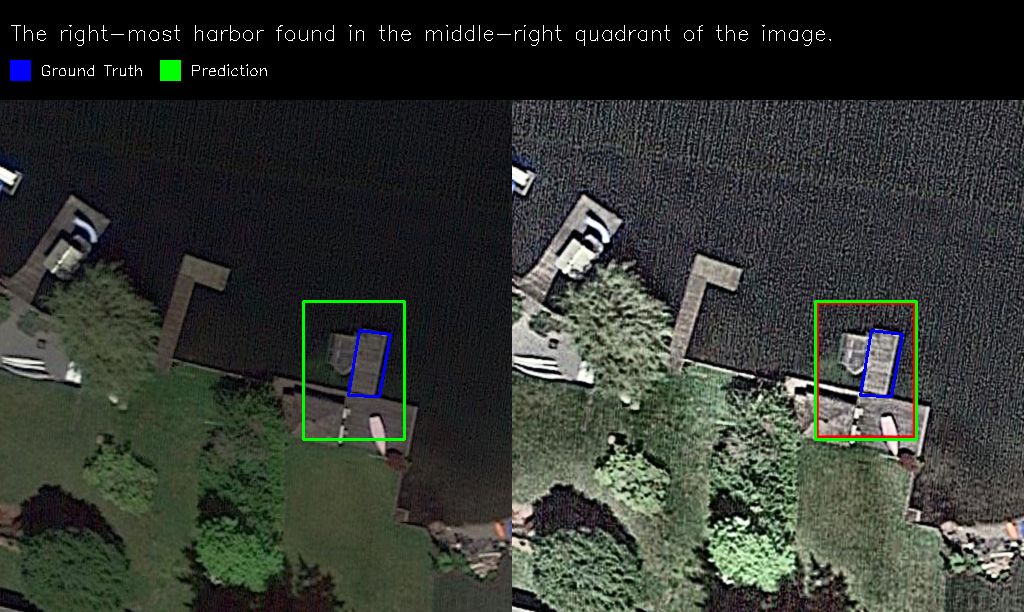}
        \caption{Proper diffusion output}
        \label{diffusion_good}
    \end{subfigure}
    \centering
    \begin{subfigure}{0.15\linewidth}
        \includegraphics[width=\linewidth]{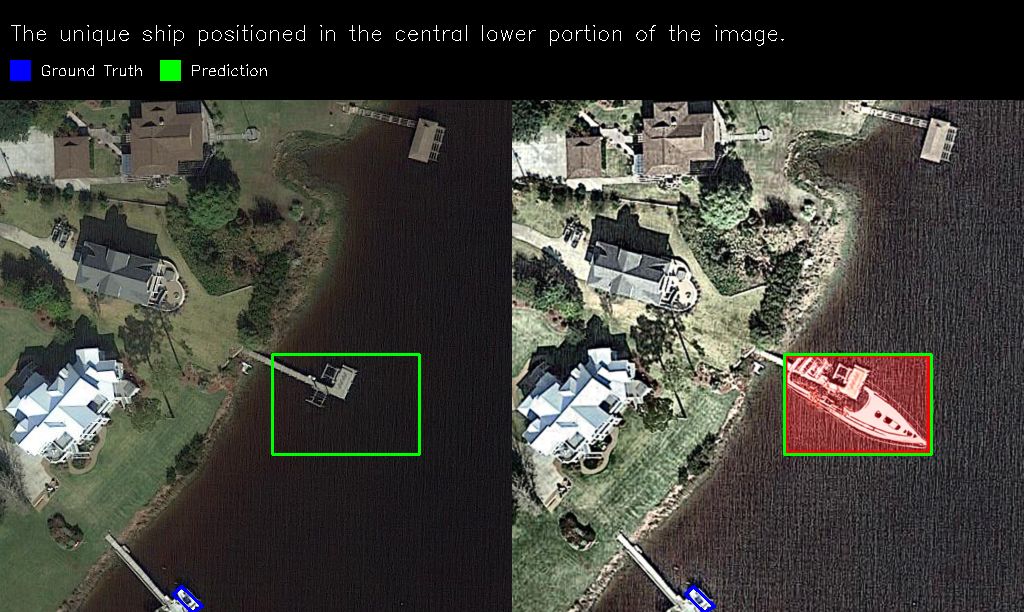}
        \caption{Diffusion hallucination}
        \label{diffusion_bad}
    \end{subfigure}
    \hfill
    \caption{Qualitative grounding results on the VRS Bench dataset. Ground-truth and predicted bounding boxes are shown in blue and green, respectively. (a) Outputs from multiple models, (b) correct diffusion-based generation, and (c) an example of diffusion hallucination caused by strong textual priors.}
    \label{grounding_results}
\end{figure*}

Figure~\ref{grounding_results} presents qualitative results from different grounding pipelines. SAM3 alone performs poorly, often not being able to detect any relevant objects in the image. The RemoteSAM-SAM3-v1 and RemoteSAM-SAM3-v2 pipelines provide noticeably improved outputs, while EarthMind performs moderately well. Subfigure~\ref{diffusion_bad} illustrates diffusion-induced hallucination, where the model generates an object based on textual cues rather than the image itself when it is unable to find the requested object in the image. In contrast, the proposed majority-voting framework effectively suppresses such errors, producing the most reliable grounding predictions.



\section{\textbf{Challenges and Future Work}}
\label{sec:challenges}

Results indicate that numeric VQA remains difficult, consistent with limitations observed in prior models. For grounding, several fixed hyperparameters—such as the 99\% DBSCAN threshold and the 10\% cropping threshold in DiffuSAM—may not generalize well across all scenarios. Future work includes developing adaptive thresholding strategies that dynamically select optimal parameters based on model confidence or image characteristics.

\section{\textbf{Conclusion}}
\label{sec:conclusion}

We introduce three new grounding pipelines—RemoteSAM-SAM3-v1, RemoteSAM-SAM3-v2, and DiffuSAM—for visual grounding in remote sensing imagery. Majority voting across existing and proposed models yields improved grounding performance. Fine-tuning the EarthMind model further enhances VQA accuracy, particularly for numeric questions. By integrating these components, we develop a unified multimodal pipeline capable of captioning, VQA, and grounding within a single framework, with optional SAR-to-optical translation. Overall, the proposed modular system provides an effective and comprehensive solution for remote-sensing scene understanding.

\section*{APIs and Licensing}
This implementation doesn't use any third-party APIs. The off the shelf open-source models used are licensed under permissive licenses such as the MIT License, SAM License and Apache-2.0 License.

\bibliographystyle{IEEEtran}
\bibliography{references}

\appendices



\section{Existing Model Architectures}
\label{appendix_models}

\begin{figure*}[t]
    \centering
    
    \begin{subfigure}{0.45\linewidth}
        \centering
        \includegraphics[ width=\linewidth]{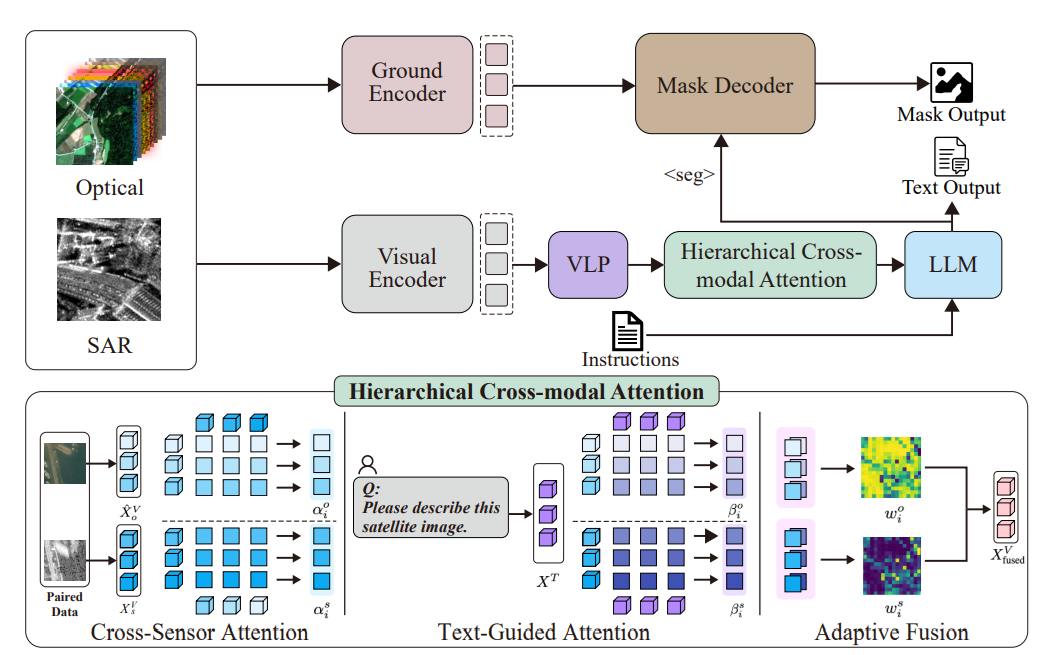}
        \caption{EarthMind architecture.}
        \label{earthmind}
    \end{subfigure}
    \hfill
    \begin{subfigure}{0.65\linewidth}
        \centering
        \includegraphics[ width=\linewidth]{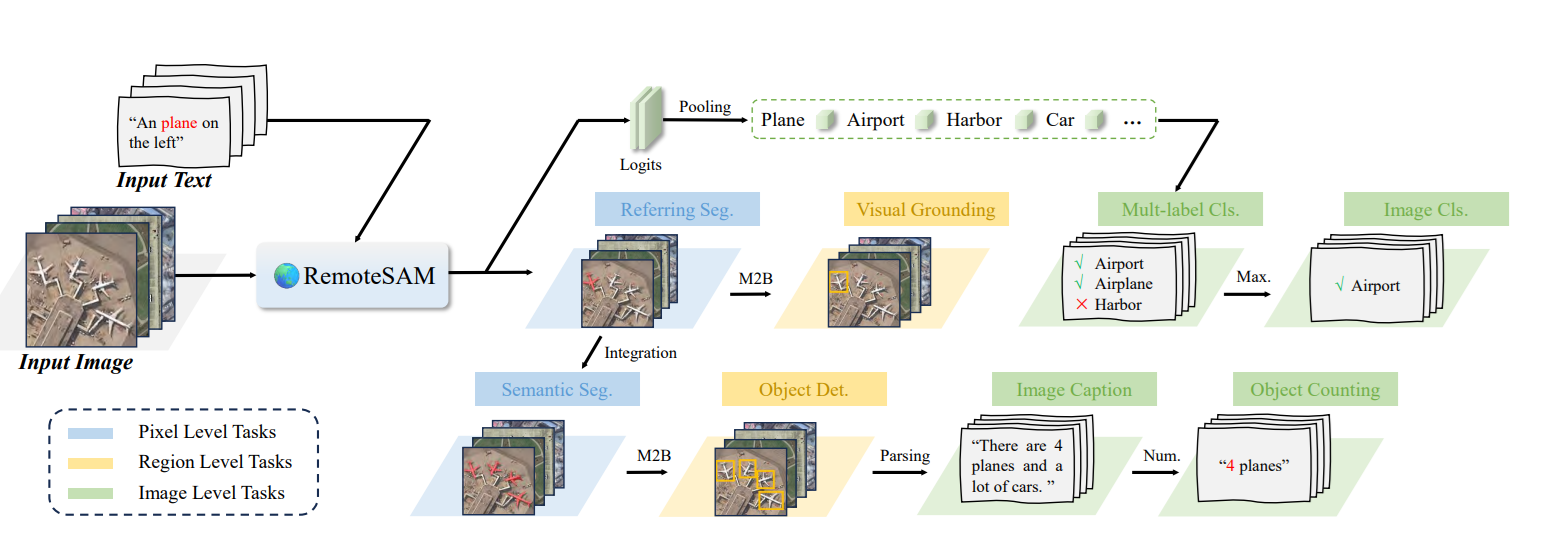}
        \caption{RemoteSAM architecture.}
        \label{remotesam}
    \end{subfigure}

    \vspace{0.3cm}

    \begin{subfigure}{0.55\linewidth}
        \centering
        \includegraphics[ width=\linewidth]{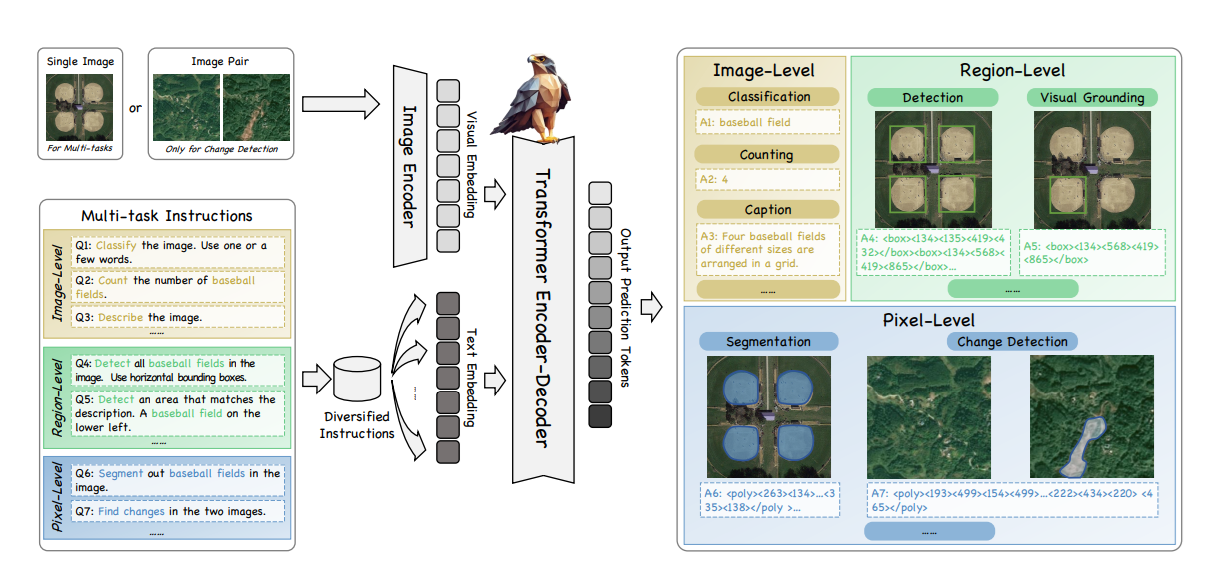}
        \caption{Falcon Architecture.}
        \label{falcon}
    \end{subfigure}
    \hfill
    \begin{subfigure}{0.3\linewidth}
        \centering
        \includegraphics[ width=\linewidth]{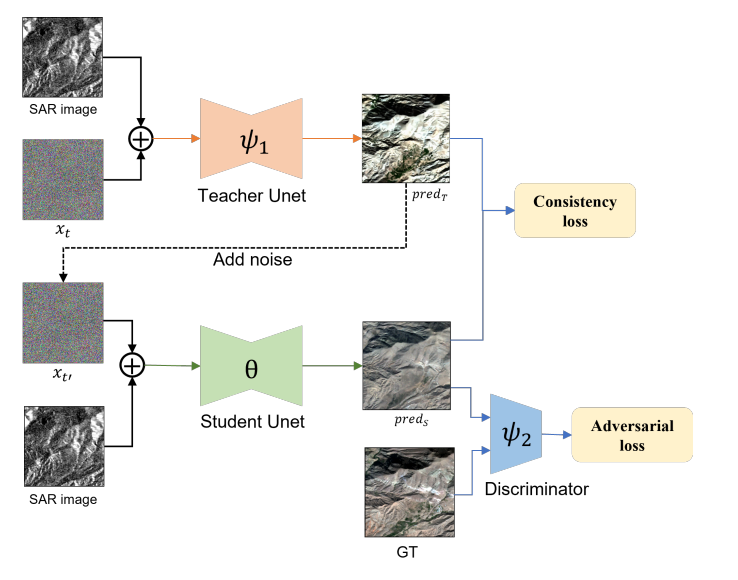}
        \caption{SAR-to-Optical Architecture.}
        \label{sar_rgb}
    \end{subfigure}

    \caption{The architectures of the components of the unified model. a) EarthMind Architecture, b) RemoteSAM Architecture, c) Falcon Architecture, d) SAR-to-Optical image translation. EarthMind is the backbone for captioning and VQA tasks, whereas majority voting output of EarthMind, RemoteSAM, SAM3, Falcon  serves for the ground model.}
    \label{architectures}
\end{figure*}

\subsection{EarthMind \cite{shu2025earthmindleveragingcrosssensordata}}


EarthMind employs two visual encoders: a global semantic encoder $E_v$ and a grounding encoder $E_g$, whose outputs are projected by a vision--language projector (VLP) into a sequence of visual tokens $X^{V}$ aligned with the LLM embedding space. These tokens, combined with text tokens, support both generative and dense prediction tasks by means of a segmentation-guidance token.

To handle heterogeneous modalities, EarthMind formats all inputs as multi-frame sequences. Single- and dual-channel SAR data are padded into pseudo-RGB frames, while multispectral bands are grouped into triplets. This unified representation promotes temporal modeling, spectral complementarity, and efficient shared-encoder processing.

EarthMind uses a hierarchical cross-modal attention which uses has two stages: a) cross-sensor attention between optical and SAR images and b) text guided attention to visual features.

\subsubsection{Cross-Sensor Attention}

To align modalities with differing resolutions, adaptive pooling of optical image yields $\hat{X}^{V}_{o} \in \mathbb{R}^{N \times D}$. Then, Bidirectional cross-modal attention is computed as:

\begin{equation}
\begin{aligned}
A^{o2s} = \mathrm{Softmax}\!\left(\frac{\hat{X}^{V}_{o} (X^{V}_s)^{T}}{\sqrt{D}}\right),\\
A^{s2o} = \mathrm{Softmax}\!\left(\frac{X^{V}_{s} (\hat{X}^{V}_{o})^{T}}{\sqrt{D}}\right).
\end{aligned}
\end{equation}

Where, $A^{o2s}_{ij}$ and $A^{s2o}_{ij}$ depict the attention from optical token i to SAR token j and SAR token i to optical token j respectively.

Modality-specific importance scores are obtained via spatial aggregation:
\begin{equation}
\alpha^{o}_{i} = \frac{1}{N}\sum_{j=1}^{N} A^{o2s}_{ij}, 
\qquad
\alpha^{s}_{i} = \frac{1}{N}\sum_{j=1}^{N} A^{s2o}_{ij}.
\end{equation}

\subsubsection{Text-Guided Attention}

Task relevance is incorporated through cross-attention between visual tokens and text embeddings. The task embeddings $X^{T} \in \mathbb{R}^{L \times D}$ is aligned with visual tokens using a projector:

\begin{equation}
\begin{aligned}
A^{o2t} = \mathrm{Softmax}\!\left(\frac{\hat{X}^{V}_{o} (X^{T})^{T}}{\sqrt{D}}\right), \\
A^{s2t}_{2} = \mathrm{Softmax}\!\left(\frac{X^{V}_{s} (X^{T})^{T}}{\sqrt{D}}\right).
\end{aligned}
\end{equation}

Text-relevance weights are computed as:
\begin{equation}
\beta^{o}_{i} = \frac{1}{L}\sum_{j=1}^{L} A^{o2t}_{ij}, 
\qquad
\beta^{s}_{i} = \frac{1}{L}\sum_{j=1}^{L} A^{s2t}_{ij}.
\end{equation}

\subsubsection{Adaptive Fusion}

The final fusion weights combine cross-modal complementarity and text relevance using a learnable balance parameter $\lambda$:
\begin{equation}
\gamma^{o}_{i} = \lambda \alpha^{o}_{i} + (1-\lambda)\beta^{o}_{i}, \qquad
\gamma^{s}_{i} = \lambda \alpha^{s}_{i} + (1-\lambda)\beta^{s}_{i}.
\end{equation}
Normalized fusion weights are obtained via
\[
[w^{o}, w^{s}] = \mathrm{Softmax}([\gamma^{o}, \gamma^{s}]),
\]
and the fused token representation is:
\begin{equation}
X^{V}_{\text{fused},i} = w^{o}_{i} \hat{X}^{V}_{o,i} + w^{s}_{i} X^{V}_{s,i}.
\label{eq:fused}
\end{equation}
This produces compact multimodal tokens $X^{V}_{\text{fused}} \in \mathbb{R}^{N \times D}$, preserving cross-sensor information while maintaining efficiency for LLM processing. Figure \ref{earthmind} presents the overall architecture of EarthMind model.

\subsubsection{Improvement of Captioning and VQA tasks}
The segmentation masks are used to highlight key regions or objects, and these mask-rendered images together with the initial captions are provided to GPT-4o to generate grounded and more detailed descriptions (RoI based summarization). This leads to improved captioning performance. The enhanced captions are expanded into diverse question–answer pairs through a few-shot self-instruct prompting strategy, producing varied and semantically rich VQA samples, thus, improving VQA performance. \textbf{This is why EarthMind performs very well in captioning and VQA tasks, but can still struggle in grounding task.}

\subsection{RemoteSAM \cite{yao2025remotesam}}
RemoteSAM is composed of three primary modules: a Visual Encoder $E_V$, a Text Encoder $E_T$, and a Mask Decoder $D_M$. The model is defined by the unified mapping from the input image $I \in \mathbb{R}^{H \times W \times 3}$ and the text prompt $T \in \mathbb{R}^{L_T}$ to the binary segmentation mask $\hat{M} \in \{0, 1\}^{H \times W}$:
$$
\text{RemoteSAM}: (I, T) \to \hat{M}
$$

The visual features $X^V \in \mathbb{R}^{N \times D}$ and text features $X^T \in \mathbb{R}^{L_T \times D}$ are extracted:
\begin{equation}
\begin{aligned}
X^V &= E_V(I) & X^T &= E_T(T).
\end{aligned}
\end{equation}
The Mask Decoder $D_M$ fuses $X^V$ and $X^T$ to produce a score map $S \in \mathbb{R}^{H \times W}$, which is then thresholded (using threshold $\tau$) to yield the binary mask $\hat{M}$.
\\
For object detection, the mask $\hat{M}$ is converted to a set of bounding boxes $B = \{b_k\}$ using the Mask-to-BBox (M2B) transformation, where $k$ is the index of the connected component $\mathcal{C}_k$:
$$
B = \bigcup_{k} \mathrm{BoundingBox}(\mathcal{C}_k(\hat{M}))
$$
This is refined by the Enhanced Processing of Object Contours (EPOC) strategy to resolve limitations in processing adjacent targets:
\begin{equation}
B_{\text{refined}} = \mathrm{Refine}_{\text{EPOC}}(B)
\end{equation}
\\
For counting objects, the count is the cardinality of the set of distinct connected components ($\mathcal{C}$) within the mask $\hat{M}$:
\begin{equation}
\text{Count} = |\mathcal{C}(\hat{M})|
\end{equation}

\subsection{SAM3 \cite{carion2025sam}}

SAM3 consists of three main components: an image encoder, a prompt encoder, and a unified detector–decoder module for producing segmentation masks. First, a high-capacity transformer-based image encoder extracts dense spatial features from the input image. The encoder is prompt-agnostic and produces a single feature map used by all downstream components. In the prompt encoder, point, box, mask, or text prompts are embedded into a set of prompt tokens. These tokens condition the segmentation process and allow flexible and user-driven control. SAM3 introduces a unified detector that operates directly on image features and optional prompt tokens. It produces instance-aware tokens representing mask proposals. These tokens remove the need for external proposal mechanisms and support class-agnostic or text-guided detection. Then, a lightweight transformer head fuses image features, prompt tokens, and detector tokens to produce final segmentation masks in a single forward pass. Although, SAM3 is trained on a massive, concept-rich dataset containing mixed images, videos, synthetic or external, but not trained specific to remote sensing dataset, it can perform worse than the other remote sensing specific models.

\subsection{Falcon \cite{yao2025falcon}}
Falcon architecture is a sequence-to-sequence framework which can utilize all the task in a generalized format. It uses a generalized dynamic prompt training strategy to avoid task specific reliance on the model. If $I$ is an remote sensing image and $T$ is a given prompt, it modifies $T$ to a predefined phrased version $T'$. For example, caption of a region can be given as `Describe the $<region>$ in the image, where $<region>$ is $<box> <x1> <y1> <x2> <y2> </box>$ representing location tokens. It can handle simultaneous multiple ($M$) instructions denoted by $\{T'_i\}^M_{i=1}$.

Next the input image $I$ and the modified prompt $T'$ are processed through image ($E_I$) and text ($E_T$) encoders to produce $E_I(I)$ and $E_T(T')$. The image features are then processed through visual adapter to produce $V$ and the combined embeddings $X=\{V,E_T(T')\}$ is fed to a transformer based encoder-decoder model ($F$) to produce the final output. Figure \ref{falcon} shows the overall architecture of Falcon model.

The overall details of the existing models such as pre-training data, model backbone, loss function and number of parameters have been summarized in table \ref{model_details}. The existing models were trained on different datasets, which can provide significant different data prior. The number of parameters also vary significantly. EarthMind also has a higher number of parameters, which can suggest the better performance compared to other models.

\begin{table*}[h]
\centering
\begin{tabular}{l p{2.5cm} p{2.5cm} p{3.5cm} c}
\hline
\textbf{Models} & \textbf{Training datasets} & \textbf{Model backbone} & \textbf{Loss function} & \textbf{Parameters (B)} \\
\hline
EarthMind & EarthMind-Bench & InternVL2\cite{chen2024expanding}, SAM2\cite{ravi2024sam2} & CE loss (text), CE + Dice loss (image) & 4 \\
\hline

RemoteSAM & RRSISD, RisBench & RMSIN \cite{liu2024rotated} & CE + Dice loss & 0.18 \\
\hline

SAM3 & SA-Co (HQ, SYN, EXT, VIDEO) & SAM1,2 \cite{kirillov2023segany, ravi2024sam2}, DETR \cite{carion2020end} & CE loss (segmentation), L1 and gIoU loss (bounding box) & 0.85 \\
\hline
Falcon & Falcon SFT & Florence-2 \cite{xiao2024florence} & CE loss (both text and image) & 0.7 \\
\hline
\end{tabular}
\caption{Model details of EarthMind, RemoteSAM, SAM3, and Falcon.}
\label{model_details}
\end{table*}

\subsection{SAR-to-Optical Image Translation \cite{bai2024accelerating}}
\label{appendix_models_sar}

This model uses a conditional diffusion process to convert SAR image to optical image. Moreover, to achieve the generation in few steps, this model uses consistency distillation training. In training, a forward diffusion process is used to add noise in the target image. The SAR image is added with the noised image as conditioning input and fed to a U-net. To utilize the distillation training, in the reverse process, a teacher U-net is used to predict the optical image. Noise is added with the predicted image using forward diffusion process. The input SAR image added with this noised input and a student U-net network is used to predict the student prediction. A consistency loss is used to align the teacher and student output. An adversarial loss is used to align the student prediction with the ground truth. Equation \ref{sar_loss} shows the loss function of this model. Figure \ref{sar_rgb} shows the overall architecture of this model.

\begin{equation}
\begin{aligned}
\mathcal{L}_{\text{consistency}}
& = \frac{1}{N} \sum_{i=1}^{N}
\left(pred^{(i)}_{T} - pred^{(i)}_{S} \right)^{2}, \\
\mathcal{L}_{G}^{\text{adv}} & = - \mathbb{E}_{x_0, c_s} \left[ D_{\psi_2}\big(f_{\theta}(x_t', c_s),\, c_s\big) \right],\\
\mathcal{L}_{D}^{\text{adv}} & =
\mathbb{E}_{x_0}\!\left[ \max\left(0,\, 1 - D_{\psi_2}\big(f_{\theta}(x_0, c_s),\, c_s\big)\right) \right]\\
& +
\mathbb{E}_{x_t}\!\left[\max\left(0,\, 1 + D_{\psi_2}\big(f_{\theta}(x_t', c_s),\, c_s\big)\right) \right],\\
\mathcal{L}_{\text{total}} & = 
\mathcal{L}_{\text{consistency}} 
+ \lambda_{\text{adv}}\, \mathcal{L}_{G}^{\text{adv}} .
\end{aligned}
\label{sar_loss}
\end{equation}

Where $x_0$ is the optical image, $c_s$ is the input SAR image, $G$ is the generator model, $f_{\theta}$ is the student model, and $D_{\psi_2}$ is the discriminator.

\section{Details of evaluation metrics}
\label{appendix_metrics}

\subsection{Captioning Metric}
\begin{itemize}
    \item \textbf{BERT-BLEU$_4$:}
Caption quality is measured using BERT-BLEU, which combines semantic similarity from BERT embeddings with BLEU-$n$ structure. For each $n$-gram level:
\begin{equation}
P_n =
\frac{1}{|R_n|} 
\sum_{r \in R_n}
\max_{c \in C_n}
\cos\!\left(
E(c), E(r)
\right),
\end{equation}
where $E(\cdot)$ denotes BERT embeddings. A length penalty is also applied:
\begin{equation}
LP =
\exp\left(
-\alpha 
\frac{|L_C - L_R|}{L_R}
\right),
\qquad \alpha = 0.5.
\end{equation}
The final caption similarity score is:
\begin{equation}
S_{captioning} = \text{BERT-BLEU}_4 =
LP \cdot 
\max_{1 \le n \le 4} P_n.
\end{equation}

\end{itemize}

\subsection{VQA Metrics}

\begin{itemize}
    \item \textbf{Binary Attribute Accuracy:} For yes/no questions, we use exact match:
\begin{equation}
S_{\mathrm{binary}} =
\begin{cases}
1, & y_{\mathrm{pred}} = y_{\mathrm{gt}}, \\
0, & \text{otherwise}.
\end{cases}
\end{equation}

\item \textbf{Numeric Attribute Score:} For numeric predictions, we evaluate using a relative error penalty:
\begin{equation}
S_{\mathrm{numeric}} =
\exp\left(
-\alpha 
\frac{|x_{\mathrm{pred}} - x_{\mathrm{gt}}|}
{x_{\mathrm{gt}}}
\right),
\qquad \alpha = 23.
\end{equation}

\item \textbf{Semantic Attribute Score:} For descriptive answers, we employ the same BERT-BLEU formulation used for captioning:
\begin{equation}
S_{\mathrm{semantic}} = 
\text{BERT-BLEU}_4.
\end{equation}

\end{itemize}

\subsection{Grounding Metrics}
\begin{itemize}
    \item \textbf{Mean Intersection over Union (mIoU):} Given a predicted bounding box $B_p$ and ground-truth box $B_g$, the Intersection over Union (IoU) is defined as:
\begin{equation}
\text{IoU}(B_p, B_g) = 
\frac{|B_p \cap B_g|}{|B_p \cup B_g|}.
\end{equation}
For $K$ matched instances, the mean IoU is computed as:
\begin{equation}
\text{mIoU} = 
\frac{1}{K} \sum_{i=1}^{K} 
\text{IoU}\big(B_p^{(i)}, B_g^{(i)}\big).
\end{equation}

\item \textbf{Accuracy@IoU\_Threshold:} A prediction is considered correct if its IoU exceeds a threshold $\tau$:
\begin{equation}
C_i =
\begin{cases}
1, & \text{if } \text{IoU}_i \ge \tau, \\
0, & \text{otherwise}.
\end{cases}
\end{equation}
Thus, accuracy at threshold $\tau$ is:
\begin{equation}
\text{Acc@}\tau = 
\frac{1}{N} \sum_{i=1}^{N} C_i.
\end{equation}


\item \textbf{Average Count Difference:}
To measure count accuracy, we compute the average absolute deviation between predicted and ground-truth object counts:
\begin{equation}
\text{ACD} = 
\frac{1}{N} \sum_{i=1}^{N}
\left|
N_{\mathrm{pred}}^{(i)} - 
N_{\mathrm{ref}}^{(i)}
\right|.
\end{equation}

\item \textbf{mIoU with Count Penalty.}
To jointly evaluate localization quality and instance count correctness, we apply an exponential penalty to the mIoU:
\begin{equation}
S_{\mathrm{grounding}} =
CP \times \text{mIoU},
\end{equation}
where the count penalty term is:
\begin{equation}
CP =
\exp\left(
-\alpha 
\left|N_{\mathrm{pred}} - N_{\mathrm{ref}}\right|
\right),
\qquad \alpha = 2.5.
\end{equation}

\end{itemize}








\end{document}